\documentclass[11pt,a4paper]{article}
\usepackage{emnlp2017}
\usepackage{times}
\usepackage{latexsym}
\usepackage{latexsym}
\usepackage{graphicx}
\usepackage{fancybox}
\usepackage{epsfig}
\usepackage{float}
\usepackage{amsmath,amsfonts,amssymb}
\usepackage{subfigure}
\usepackage{multirow}
\usepackage{bm}
\usepackage[]{algorithm2e}
\usepackage{color}
\usepackage{hyperref}
\usepackage{tikz}
\usepackage{wasysym}
\definecolor{mygreen}{rgb}{0,0.5,0.2}

\usepackage{amsmath,amsfonts,amssymb,bm}

\usepackage{xcolor}
\usepackage{tcolorbox}

\newcommand{\tableref}[1]{Table \ref{#1}}

\renewcommand{\vec}[1]{\mathbf{#1}}

\newcommand{\mat}[1]{\mathbf{#1}}
\newcommand{\trans}[1]{#1^{\textsf{T}}}

\newcommand{\argmax}[1]{\text{arg}\max_{#1}}

\usepackage{adjustbox}
\usepackage{array}
\usepackage{booktabs}

\newcolumntype{R}[2]{%
    >{\adjustbox{angle=#1,lap=\width-(#2)}\bgroup}%
    l%
    <{\egroup}%
}

\newif{\ifhidecomments}
\hidecommentsfalse

\ifhidecomments
    \newcommand{\yfcomment}[1]{}
    \newcommand{\chenhao}[1]{}
    \newcommand{\smcomment}[1]{}
    \newcommand{\yccomment}[1]{}
    \newcommand{\todo}[1]{}
    \newcommand{\nascomment}[1]{}
\else
    \newcommand{\nascomment}[1]{\textcolor{blue}{[\textsc{#1  ---nas}]}}
    \newcommand{\yfcomment}[1]{[\textcolor{blue}{\tt #1 - YJ}]}
    \newcommand{\chenhao}[1]{[\textcolor{brown}{\tt #1 - CT}]}
    \newcommand{\smcomment}[1]{[\textcolor{mygreen}{\tt #1 - SM}]}
    \newcommand{\yccomment}[1]{[\textcolor{cyan}{\tt #1 - YC}]}
    \newcommand{\todo}[1]{[\textcolor{red}{\tt TODO for YJ: #1}]}
\fi

\renewcommand{\trans}[1]{\ensuremath{#1^{\top}}}

\usepackage[normalem]{ulem}

\emnlpfinalcopy

\def\emnlppaperid{1076}

\newcommand{\entone}[1]{[\emph{\color{blue} #1}]$_1$}
\newcommand{\enttwo}[1]{[\emph{\color{red} #1}]$_2$}
\newcommand{\entthree}[1]{[\emph{\color{mygreen} #1}]$_3$}
\newcommand{\entfour}[1]{[\emph{\color{violet} #1}]$_4$}
\newcommand{\entfive}[1]{[\emph{\color{orange} #1}]$_5$}
\newcommand{\entsix}[1]{[\emph{\color{olive} #1}]$_5$}
\newcommand{\model}{{\sc EntityNLM}\xspace}
\newcommand{\new}[1]{#1}

\title{Dynamic Entity Representations in Neural Language Models}

\author{
  Yangfeng Ji$^{\ast}$\quad Chenhao Tan$^{\ast}$\quad Sebastian
  Martschat$^{\dagger}$ \quad
  {\bf Yejin Choi$^{\ast}$\quad Noah A. Smith$^{\ast}$}\\
  $^\ast$Paul G. Allen School of Computer Science \& Engineering, University of Washington \\ 
  $^\dagger$Department of Computational Linguistics, Heidelberg University \\ 
  {\tt \{yangfeng,chenhao,yejin,nasmith\}@cs.washington.edu}\\
  {\tt martschat@cl.uni-heidelberg.de}
}

\date{}

\begin{document}

\maketitle

\begin{abstract}
%
  Understanding a long document 
  requires tracking how entities are introduced and evolve over time.
  We present a new type of language model, \model, that can explicitly model entities, 
  dynamically update their representations, and contextually generate their mentions. 
  Our model 
  is generative and flexible; it can model an arbitrary number of entities in context 
  while generating each entity mention at an arbitrary length.  
  In addition, it can be used for several different tasks such as language modeling, coreference resolution, and entity prediction.
  Experimental results with all these tasks demonstrate that our model consistently outperforms strong baselines and prior work. 
\end{abstract}


\section{Introduction}
\label{sec:intro}

Understanding a narrative requires keeping track of its participants over a long-term context. 
As a story unfolds, the information a reader associates with each character in a story increases, and expectations about what will happen next change accordingly.  At present, models of natural language do not explicitly track entities; indeed, in today's language models, entities are no more than the words used to mention them.

In this paper, we endow a generative language model with the ability to build up a dynamic representation of each entity mentioned in the text.  Our language model defines a probability distribution over the whole text, 
with a distinct generative story for entity mentions. 
It explicitly groups those mentions that corefer and associates with each entity a continuous representation that is updated by every contextualized mention of the entity, and that in turn affects the text that follows.

Our method builds on 
recent 
advances in representation learning, creating local probability distributions from neural networks.  It can be understood as a recurrent neural network language model, augmented with random variables for entity mentions that capture coreference, and with dynamic representations of entities. 
We estimate the model's parameters from data that is annotated with entity mentions and coreference.

\begin{figure}
  \centering
  \begin{tabular}{p{0.45\textwidth}} 
    \toprule
    \entone{John} wanted to go to \enttwo{the coffee shop} in \entthree{downtown Copenhagen}. \entone{He} was told that \enttwo{it} sold  \entfour{the best beans}. \\
    \bottomrule
  \end{tabular}
  \caption{\model explicitly tracks entities in a text, including coreferring relationships between entities like \entone{John} and \entone{He}.  As a language model, it is designed to predict that a coreferent of \enttwo{the coffee shop} is likely to follow ``\emph{told that},'' that the referring expression will be ``\emph{it}'', 
  and that ``\emph{sold the best beans}'' is likely to come next,
  by using entity information encoded in the dynamic distributed representation.
    \label{fig:example}}
\end{figure}

Because our model is generative, it can be queried in different ways.  Marginalizing everything except the words, it can play the role of a language model.  
In \S\ref{subsec:app-lm}, we find that it outperforms both a strong $n$-gram language model and a strong recurrent neural network language model on the English test set of the CoNLL 2012 shared task on coreference evaluation~\citep{pradhan2012conll}. 
The model can also identify entity mentions and coreference relationships among them. 
In \S\ref{subsec:app-cr}, we show that it can easily be used to add a performance boost to a strong coreference resolution system, by reranking a list of $k$-best candidate outputs.
On the CoNLL 2012 shared task test set, the reranked outputs are significantly better than the original top choices from the same system.
Finally, the model can perform entity cloze tasks.  
As presented in \S\ref{subsec:app-ep}, it achieves state-of-the-art performance on the InScript corpus~\citep{modi2017modeling}.

\section{Model}
\label{sec:model}

A language model defines a distribution over sequences of word tokens;
let $X_t$ denote the random variable for the $t$th word in the
sequence, $x_t$ denote the value of $X_t$ and $\vec{x}_t$ the distributed 
representation (embedding) of this word.  
Our starting point for language modeling is a recurrent neural network 
\citep{mikolov2010recurrent}, which defines
\begin{align}
  p(X_t \mid \text{history}) & = \mathrm{softmax} \left(\mat{W}_h
                               \vec{h}_{t-1} + \vec{b}\right)
                               \label{eq:rnnlm}\\
  \vec{h}_{t-1} &= \text{\sc lstm}(\vec{h}_{t-2}, \vec{x}_{t-1}) 
                  \label{eq:hidden-state}
\end{align}
where $\mat{W}_h$ and $\vec{b}$ are parameters of the model (along
with word embeddings $\vec{x}_t$), \textsc{lstm} is the widely used recurrent function known as
``long short-term memory'' \citep{hochreiter1997long}, 
and $\vec{h}_t$ is a \textsc{lstm} hidden state encoding 
the history of the sequence up to the $t$th word.

Great success has been reported for this model \citep{zaremba2015recurrent}, which
posits nothing explicitly about the words appearing in the text
sequence.  Its generative story is simple:  the value of each $X_t$ is randomly
chosen conditioned on the vector $\vec{h}_{t-1}$ encoding its history.

\newcommand{\newentitysym}{0\xspace}
\newcommand{\notentitysym}{\clock\xspace}

\subsection{Additional random variables and representations for entities}

To introduce our model, we associate with each word an additional set of random variables.  
At position $t$, 
\begin{itemize}
\item $R_t$ is a binary random variable that indicates whether $x_t$ 
  belongs to an entity mention ($R_t=1$) or not ($R_t = 0$).  Though
  not explored here, this is easily generalized to a categorial
  variable for the \emph{type} of the entity (e.g., person,
  organization, etc.).
\item $L_t\in\{1, \dots, \ell_{\mathit{max}}\}$ is a categorical random variable if $R_t = 1$, 
  which indicates the number of remaining words in this mention,
  including the current word (i.e., $L_t = 1$ for the last word in any mention). 
  $\ell_{\mathit{max}}$ is a predefined maximum length fixed to be 25, 
  which is an empirical value derived from the training corpora used in the experiments.
  If $R_t=0$, then $L_t = 1$. 
  We denote the value of $L_t$ by~$\ell_t$. 
\item $E_t\in\mathcal{E}_t$ is the index of the entity referred to, if $R_t=1$.  
  The set $\mathcal{E}_t$ consists of $\{1, \ldots, 1 + \max_{t' <
    t} e_{t'}\}$, i.e., the indices of all previously mentioned entities
  plus an additional value for a new entity.  Thus $\mathcal{E}_t$
  starts as $\{1\}$ and
  grows monotonically with $t$, allowing for an arbitrary number of
  entities to be mentioned. We denote the value of $E_t$ by $e_t$.  
  If $R_t = 0$, then $E_t$ is fixed to a special value $\clock$.
\end{itemize}
The values of these random variables for our running example are shown in~\autoref{example2}.

In addition to using symbolic variables to encode mentions
and coreference relationships, we maintain a vector representation of
each entity that evolves over time.  For the $i$th entity,
let $\vec{e}_{i,t}$ be its representation at time $t$.  These
vectors are different from word vectors ($\vec{x}_t$), 
in that they are not parameters of the model.  They are
similar to history representations ($\vec{h}_t$), in that they are
derived through parameterized functions of the random variables' values, which we will describe in the next subsections.

\begin{figure*}
  \small 
  \begin{center}
    \begin{tabular}{rcccccccccccc}
      \toprule
      $X_{1:12}$: & {John} &wanted& to &go &to& the &coffee& shop&in & downtown &Copenhagen& . \\ 
      $R_{1:12}$: & 1 & 0 & 0 & 0 & 0 & 1 & 1 & 1 & 0 & 1 & 1 & 0 \\ 
      $E_{1:12}$: & 1 & $\notentitysym$ & $\notentitysym$ & $\notentitysym$ & $\notentitysym$ & 2 & 2 & 2 & $\notentitysym$ & 3 & 3 & $\notentitysym$ \\ 
      $L_{1:12}$: & 1 & 1 & 1 & 1 & 1 & 3 & 2 & 1 & 1 & 2 & 1 & 1 \\ 
      \hline
    \end{tabular}
    
    \begin{tabular}{rccccccccccccc}
      $X_{13:22}$: & He & was& told  &that & it & sold & the & best & beans & . \\ 
      $R_{13:22}$: & 1 & 0 & 0 & 0 & 1 & 0 & 1 & 1 & 1 & . \\ 
      $E_{13:22}$: & 1 & $\notentitysym$ & $\notentitysym$ & $\notentitysym$ & 2 & $\notentitysym$ & 4 & 4 & 4 & $\notentitysym$ \\
      $L_{13:22}$: & 1 & 1 & 1 & 1 & 1 & 1 & 3 & 2 & 1 & 0 \\
      \bottomrule
    \end{tabular}
  \end{center}
  \caption{The random variable values in \model for the running
    example in \autoref{fig:example}. \label{example2}
}
\end{figure*}

\subsection{Generative story}
\label{subsec:generative}

The generative story for the word (and other variables) at timestep
$t$ is as follows; forward-referenced equations are in the detailed
discussion that follows.
\begin{enumerate}
\item If $\ell_{t-1} = 1$ (i.e., $x_t$ is \emph{not} continuing an
  already-started entity mention):
  \begin{itemize}
  \item Choose $r_t$ (\autoref{eq:r}).
  \item If $r_t = 0$, set $\ell_t=1$ and $e_t=\notentitysym$; then go to step 3.
    Otherwise: 
    \begin{itemize}
      \item If there is no embedding for the new candidate entity with
        index $1 + \max_{t'<t} e_{t'}$, create one following~\S\ref{sec:entity-vector}.
    \item Select the entity $e_t$ from $\{1,\dots,1 + \max_{t'<t}
      e_{t'}\}$ (\autoref{eq:e}). 
    \item Set $\vec{e}_{\mathit{current}} = \vec{e}_{e_t,t-1}$, 
      which is the entity embedding of $e_t$ before timestep $t$.
    \item Select the length of the mention, $\ell_t$ (\autoref{eq:l}).
    \end{itemize}
  \end{itemize}
\item Otherwise, 
  \begin{itemize}
  \item Set $\ell_t = \ell_{t-1} - 1$, $r_t = r_{t-1}$, $e_t = e_{t-1}$.
  \end{itemize}
\item Sample $x_t$ from the word distribution given the LSTM hidden state
  $\vec{h}_{t-1}$ and the current (or most recent) entity embedding
  $\vec{e}_{\mathit{current}}$ (\autoref{eq:x}). (If $r_t=0$, then
  $\vec{e}_{\mathit{current}}$ still represents the most recently
  mentioned entity.) 
\item Advance the RNN, i.e., feed it the word vector $\vec{x}_t$ to compute $\vec{h}_t$ (\autoref{eq:hidden-state}).
\item If $r_t =1$, 
update $\vec{e}_{e_t,t}$ using $\vec{e}_{e_t,t-1}$ and $\vec{h}_t$, then  set $\vec{e}_{\mathit{current}}=\vec{e}_{e_t,t}$. 
    Details of the entity updating are given in \S\ref{sec:entity-vector}.

  \item For every entity \new{$e_{\iota} \in\mathcal{E}_t \setminus \{e_t\}$}, set
    $\vec{e}_{\iota,t} = \vec{e}_{\iota,t-1}$
    (i.e., no changes to other entities'
    representations).
  \end{enumerate}
Note that at any given time step $t$, $\vec{e}_{\mathit{current}}$ will always
contain the most recent vector representation of the most recently
mentioned entity. 

A generative model with a similar hierarchical structure was used by
\citet{haghighi2010coreference} for coreference resolution.  Our
approach differs in two important ways.  First, our model defines
a joint distribution over all of the text, not just the entity
mentions.  Second, we use representation learning rather than Bayesian
nonparametrics, allowing natural integration with the language model.

\subsection{Probability distributions}
\label{subsec:prob-dist}

The generative story above referenced several parametric distributions
defined based on vector representations of histories and entities.  
These are defined as follows.

For $r \in \{0,1\}$,
\begin{align}
  \label{eq:r}
  p(R_t = r  \mid \vec{h}_{t-1}) \propto \exp  (\trans{\vec{h}_{t-1}}\mat{W}_r\vec{r}),
\end{align}
where $\vec{r}$ is the parameterized embedding associated with $r$, which paves the way for exploring entity type representations in future work; \new{$\mat{W}_r$ is a parameter matrix for the bilinear score for $\vec{h}_{t-1}$ and $\vec{r}$}.

To give the possibility of predicting a new entity, 
we need an entity embedding beforehand with index $(1 + \max_{t'<t} e_{t'})$, which is randomly sampled from~\autoref{eq:e-init}. 
Then, for every $e \in \{1, \dots, 1 + \max_{t'<t} e_{t'}\}$: 
\begin{equation}
  \label{eq:e}
  \begin{split}
    p&(E_t = e  \mid R_t=1, \vec{h}_{t-1}) \\
    & \propto \exp(\trans{\vec{h}_{t-1}}\mat{W}_{\mathit{entity}}\vec{e}_{e,t-1}
    +\vec{w}_{\mathit{dist}}^\top
    \vec{f}(e)),
  \end{split}
\end{equation}
where $\vec{e}_{e,t-1}$ is the embedding of entity $e$ at time step
$t-1$ and $\mat{W}_{\mathit{entity}}$ is the weight matrix for
predicting entities using their continuous representations.
The score above is normalized over values $\{1, \ldots, 1 + \max_{t'<t} e_{t'}\}$. 
$\vec{f}(e)$ represents a vector of distance features associated with $e$ and the mentions of the existing entities.
\new{
  Hence two information sources are used to predict the next entity:
  (i) contextual information $\vec{h}_{t-1}$, and (ii) distance
  features $\vec{f}(e)$ from the current mention to the closest
  mention from each previously mentioned entity.
  $\vec{f}(e)=\vec{0}$ if $e$ is a new entity.}
This term can also be extended to include other surface-form features for coreference resolution~\citep{martschat2015latent,clark2016improving}.

For the chosen entity $e_t$ from \autoref{eq:e}, the distribution over
its mention length is drawn according to
\begin{equation}
  \label{eq:l} 
  \begin{split}
    p(L_t = \ell & \mid \vec{h}_{t-1}, \vec{e}_{e_t, t-1}) \\
    & \propto \exp
    (\new{\trans{\mat{W}_{\mathit{length},\ell}}}[\vec{h}_{t-1}; \vec{e}_{e_t, t-1}]), 
  \end{split}
\end{equation}
where $\vec{e}_{e_t,t-1}$ is the most recent embedding of the entity $e_t$, not updated with $\vec{h}_t$.
\new{The intuition is that $\vec{e}_{e_t,t-1}$ will help contextual information $\vec{h}_{t-1}$ to select the residual length of entity $e_t$.}
$\mat{W}_{length}$ is the weight matrix for length prediction, with $\ell_{\mathit{max}}=25$ rows.

Finally, the probability of a word $x$ as the next token is \new{jointly modeled by $\vec{h}_{t-1}$ and the vector representation of the most recently mentioned entity $\vec{e}_{\mathit{current}}$}:
\begin{equation}
  \label{eq:x}
  \begin{split}
    p(X_t & = x \mid \vec{h}_{t-1}, \vec{e}_{\mathit{current}}) \\
    & \propto \textsc{cfsm}(\vec{h}_{t-1} + \mat{W}_{e}\vec{e}_{\mathit{current}}),
  \end{split}
\end{equation}
where $\mat{W}_{e}$ is a transformation matrix to adjust the dimensionality of $\vec{e}_{\mathit{current}}$.
{\sc cfsm} is a class factorized softmax function~\citep{goodman2001classes,baltescu2015pragmatic}.
It uses a two-step prediction with predefined word classes instead of direct prediction on the whole vocabulary, and reduces the time complexity to the log of vocabulary size.

\subsection{Dynamic entity representations}
\label{sec:entity-vector}

Before predicting the entity at step $t$, we need an embedding for 
the new candidate entity with index $e'=1+\max_{t'<t}e_{t'}$ if it does not exist. 
The new embedding is generated randomly, according to a normal distribution, 
then projected onto the unit ball:
\begin{equation}
  \label{eq:e-init}
  \begin{split}
    \vec{u} & \sim \mathcal{N}(\vec{r}_1, \sigma^2 \mat{I}); \\
    \vec{e}_{e', t-1} & = \frac{\vec{u}}{\|\vec{u}\|_2},
  \end{split}
\end{equation}
where $\sigma=0.01$. The time step $t-1$ in $\vec{e}_{e', t-1}$ means the current embedding contains no information from step $t$,
although it will be updated once we have $\vec{h}_t$ and if $E_t=e'$.
$\vec{r}_1$ is the parameterized embedding for $R_t=1$, \new{which will be jointly optimized with other parameters and is expected to encode some generic information about entities}.
All the initial entity embeddings are centered on the mean $\vec{r}_1$, which is used in~\autoref{eq:r} to determine whether the next token belongs to an entity mention.
Another choice would be to initialize with a zero vector, although our preliminary experiments showed this did not work as well as random initialization in~\autoref{eq:e-init}.


Assume $R_t=1$ and $E_t=e_t$, which means $x_t$ is part of a mention of entity $e_t$.
Then, we need to update $\vec{e}_{e_t,t-1}$ based on the new information we have from $\vec{h}_t$. 
The new embedding $\vec{e}_{e_t,t}$ is a convex combination of the old embedding
($\vec{e}_{e_t,t-1}$) and current LSTM hidden state ($\vec{h}_{t}$) with
the interpolation ($\delta_t$) determined dynamically based on a bilinear function:
\begin{eqnarray}
  \label{eq:entity-update}
    \delta_t & = & \sigma(\vec{h}_{t}^\top \mat{W}_\delta\vec{e}_{e_t,t-1}); \nonumber \\
    \vec{u} & = & \delta_t \vec{e}_{e_t,t-1} + (1-\delta_t) \vec{h}_{t}; \nonumber \\
    \vec{e}_{e_t,t} & = & \frac{\vec{u}}{\|\vec{u}\|_2}, 
\end{eqnarray}
This updating scheme will be used to update $e_t$ in \emph{each} of all the following $\ell_t$ steps.
The projection in the last step keeps the magnitude of the entity
embedding fixed, avoiding numeric overflow.
A similar updating scheme has been used by \citet{henaff2016tracking}
for the ``memory blocks'' in their recurrent entity network models.
The difference is that their model updates all memory blocks in each
time step.
Instead, our updating scheme in~\autoref{eq:entity-update} only applies to the selected 
entity $e_t$ at time step $t$.




\subsection{Training objective}

The model is trained to maximize the log of the joint probability of $\vec{R}, \vec{E}, \vec{L}$, and $\vec{X}$:
\begin{equation}
  \label{eq:objective}
  \begin{split}
    \ell(\bm{\theta}) &= \log P(\vec{R},\vec{E},\vec{L},\vec{X}; \bm{\theta})\\
    &=\sum_{t}\log P(R_t, E_t, L_t, X_t; \bm{\theta}),
  \end{split}
\end{equation}
where $\bm{\theta}$ is the collection of all the parameters in this model. 
Based on the formulation in~\S\ref{subsec:prob-dist}, \autoref{eq:objective} 
can be decomposed as the sum of conditional log-probabilities of each random 
variable at each time step.  

This objective requires the training data annotated as in \autoref{example2}. 
We do not assume that these variables are observed at test time.






\section{Implementation Details}
\label{sec:imp}
Our model is implemented with DyNet \citep{neubig2017dynet} \new{and available at \url{https://github.com/jiyfeng/entitynlm}}.
We use AdaGrad~\citep{duchi2011adaptive} with learning rate $\lambda=0.1$ and ADAM~\citep{kingma2014adam} with default learning rate $\lambda=0.001$ as the candidate optimizers of our model. 
For all the parameters, we use the initialization tricks recommended by \citet{glorot2010understanding}.
To avoid overfitting, we also employ dropout~\citep{srivastava2014dropout} with the candidate rates as $\{0.2, 0.5\}$. 

In addition, there are two tunable hyperparameters of \model: the size of word embeddings and the dimension of LSTM hidden states.
For both of them, we consider the values $\{32, 48, 64, 128, 256\}$. 
We also experiment with the option to either use the pretrained GloVe word embeddings~\citep{pennington2014glove} or randomly initialized word embeddings (then updated during training).
For all experiments, the best configuration of hyperparameters and optimizers is selected based on the objective value on the development data.


\section{Evaluation Tasks and Datasets}
\label{sec:datasets}


We evaluate our model in diverse use scenarios: (i) language modeling, (ii) coreference resolution, and (iii) entity prediction.
The evaluation on language modeling shows how the internal entity representation, when marginalized out, can improve the perplexity of language models. The evaluation on coreference resolution experiment shows how our new language model can improve a competitive coreference resolution system. Finally, we employ an entity cloze task 
to demonstrate the generative performance of our model in predicting the next entity given the previous context.

We use two datasets for the three evaluation tasks. For language modeling and coreference resolution, we use the English benchmark data from the CoNLL 2012 shared task on coreference resolution~\citep{pradhan2012conll}. 
We employ the standard training/development/test split, which includes 2,802/343/348 documents with roughly 1M/150K/150K tokens, respectively.
We follow the coreference annotation in the CoNLL dataset to extract entities and ignore the singleton mentions in texts.

For entity prediction, we employ the InScript corpus created by \citet{modi2017modeling}.
It consists of 10 scenarios, including grocery shopping, taking a flight, etc.
It includes 910 crowdsourced simple narrative texts in total and 18 stories were ignored due to labeling problems~\citep{modi2017modeling}.
On average, each story has 12.4 sentences, 24.9 entities and 217.2 tokens.
Each 
entity mention is labeled with its entity index.
We use the same training/development/test split as in \citep{modi2017modeling}, which includes 619, 91, 182 texts, respectively.


\subsection*{Data preprocessing}
For the CoNLL dataset, we lowercase all tokens and remove any token that only contains a punctuation symbol unless it is in an entity mention. 
We also replace numbers in the documents with the special token {\sc num} and low-frequency word types with {\sc unk}. 
The vocabulary size of the CoNLL data after preprocessing is 10K.
For entity mention extraction, in the CoNLL dataset, one entity mention could be embedded in another. 
For embedded mentions, only the enclosing entity mention is kept.
We use the same preprocessed data for both language modeling and coreference resolution evaluation.

For the InScript corpus, we apply similar data preprocessing to lowercase all tokens, and we replace low-frequency word types with {\sc unk}.
The vocabulary size after preprocessing is 1K.




\section{Experiments}


In this section, we present the experimental results on the three evaluation tasks.


\subsection{Language modeling}
\label{subsec:app-lm}


\paragraph{Task description.} 
The goal of language modeling is to compute the marginal probability:
\begin{equation}
  P(\vec{X})=\sum_{\vec{R}, \vec{E}, \vec{L}}P(\vec{X}, \vec{R}, \vec{E}, \vec{L}).
\end{equation} 
However, due to the long-range dependency in recurrent neural networks, the search space of 
$\vec{R}, \vec{E}, \vec{L}$ during inference grows exponentially.
We thus use importance sampling to approximate the marginal distribution of $\vec{X}$.
Specifically, with the samples from a proposal distribution $Q(\vec{R},\vec{E},\vec{L}|\vec{X})$, the approximated marginal probability is defined as
\begin{equation}
  \label{eq:marginal}
  \begin{split}
    P(&\vec{X}) = \sum_{\vec{R}, \vec{E}, \vec{L}}P(\vec{X}, \vec{R}, \vec{E}, \vec{L})\\
    &= \sum_{\vec{R}, \vec{E}, \vec{L}} Q(\vec{R},\vec{E},\vec{L}\mid \vec{X})\frac{P(\vec{X}, \vec{R}, \vec{E}, \vec{L})}{Q(\vec{R},\vec{E},\vec{L}\mid \vec{X})}\\
    & \approx \frac{1}{N}\sum_{\{\vec{r}^{(i)},\vec{e}^{(i)},\vec{\ell}^{(i)}\}\sim Q}\frac{P(\vec{r}^{(i)},\vec{e}^{(i)},\vec{\ell}^{(i)}, \vec{x})}{Q(\vec{r}^{(i)},\vec{e}^{(i)},\vec{\ell}^{(i)}\mid \vec{x})}
  \end{split}
\end{equation}
A similar idea of using importance sampling for language modeling evaluation has been used by \citet{dyer2016recurrent}.

For language modeling evaluation, we train our model on the training set from the CoNLL 2012 dataset with coreference annotation. 
On the test data, we treat coreference structure as latent variables and use importance sampling to approximate the marginal distribution of $\vec{X}$.
For each document, the model randomly draws $N=100$ samples from the proposal distribution, discussed next.

\paragraph{Proposal distribution.}
For implementation of $Q$, we use a discriminative variant of \model by taking the current word $x_t$ for predicting the entity-related variables in the same time step.
Specifically, in the generative story described in \S\ref{subsec:generative}, we delete step 3 (words are not generated, but rather conditioned upon), move step 4 before step 1, and replace $\vec{h}_{t-1}$ with $\vec{h}_t$ in the steps for predicting entity type $R_t$, entity $E_t$ and mention length $L_t$.
This model variant provides a conditional probability $Q(R_t,E_t,L_t \mid X_t)$ at each timestep.

\paragraph{Baselines.} We compare the language modeling performance with two competitive baselines: 5-gram language model implemented in KenLM \citep{heafield2013scalable} and RNNLM with LSTM units implemented in  DyNet~\citep{neubig2017dynet}. For RNNLM, we use the same hyperparameters described in~\S\ref{sec:imp} and grid search on the development data to find the best configuration. 

\paragraph{Results.} 
The results of \model and the baselines on both development and test data are reported in \autoref{tab:ppl}.
For \model, we use the value of $2^{-\frac{1}{T}\sum_{t=1}^{T}\log P(X_t,R_t,E_t,L_t)}$ on the development set with coreference annotation to select the best model configuration and report the best number.
On the test data, we are able to calculate perplexity by marginalizing all other random variables using \autoref{eq:marginal}.
To compute the perplexity numbers on the test data, our model only takes account of log probabilities on word prediction.
The difference is that coreference information is only used for training \model and not for test.
All three models reported in~\autoref{tab:ppl} share the same vocabulary, therefore the numbers on the test data are directly comparable.
As shown in \autoref{tab:ppl}, \model outperforms both the 5-gram language model and the RNNLM on the test data.
Better performance of \model on language modeling can be expected, if we also use the marginalization method defined in \autoref{eq:marginal} on the development data to select the best configuration.
However, we plan to use the same experimental setup for all experiments, instead of customizing our model for each individual task.

\begin{table}
  \centering
  \begin{tabular}{p{0.3\textwidth}l}
    \toprule
    Model & Perplexity\\
    \midrule
    1. 5-gram LM & 138.37 \\
    2. RNNLM & 134.79 \\
    3. \model & {\bf 131.64}\\
    \bottomrule
  \end{tabular}
  \caption{Language modeling evaluation on the test
    sets of the English section in the CoNLL 2012 shared task. 
    As mentioned in~\S\ref{sec:datasets}, the vocabulary size is 10K. 
    \model does not require any coreference annotation on the test data. 
  }
  \label{tab:ppl}
\end{table}


\subsection{Coreference reranking}
\label{subsec:app-cr}

\paragraph{Task description.}
We show how \textsc{EntityLM}, which allows an efficient computation
of the probability $P(\vec{R},\vec{E},\vec{L},\vec{X})$,  can be used
as a coreference reranker to improve a competitive coreference
resolution system due to \citet{martschat2015latent}.
This task is analogous to the reranking approach used in machine translation~\citep{shen2004discriminative}. 
The specific formulation is as follows:
\begin{equation}
  \label{eq:coref-resolution}
  \argmax{\{\vec{r}^{(i)},\vec{e}^{(i)},\vec{l}^{(i)}\}\in\mathcal{K}} P(\vec{r}^{(i)},\vec{e}^{(i)},\vec{l}^{(i)}, \vec{x})
\end{equation}
where $\mathcal{K}$ is the $k$-best list for a given document. In our experiments, $k = 100$. 
To the best of our knowledge, the problem of obtaining $k$-best outputs of a coreference resolution system has not been studied before.

\paragraph{Approximate $k$-best decoding.}

We rerank the output of a system that predicts an antecedent for each mention by relying on pairwise scores for mention pairs. 
This is the dominant approach for coreference resolution~\citep{martschat2015latent,clark2016ranking}. 
The predictions induce an antecedent tree, which represents antecedent decisions for all mentions in the document. 
Coreference chains are obtained by transitive closure over the antecedent decisions encoded in the tree.
A mention also can have an empty mention as antecedent, which denotes that the mention is non-anaphoric.

For extending Martschat and Strube's greedy decoding approach to $k$-best inference, we cannot simply take the $k$ highest scoring trees according to the sum of edge scores, because different trees may represent the same coreference chain. 
Instead, we use an heuristic that creates an approximate $k$-best list on candidate antecedent trees. 
The idea is to generate trees from the original system output by considering suboptimal antecedent choices that lead to different coreference chains. 
For each mention pair $(m_j, m_i)$, we compute the difference of its score to the score of the optimal antecedent choice for $m_j$. 
We then sort pairs in ascending order according to this difference and iterate through the list of pairs. 
For each pair $(m_j,m_i)$, we create a tree $t_{j,i}$ by replacing the antecedent of $m_j$ in the original system output with $m_i$. 
If this yields a tree that encodes different coreference chains from all chains encoded by trees in the $k$-best list, we add $t_{i,j}$ to the $k$-best list.
In the case that we cannot generate a given number of trees (particularly for a short document with a large $k$), we pad the list with the last item added to the list.

\paragraph{Evaluation measures.} For coreference resolution evaluation, we employ the CoNLL scorer \citep{pradhan2014scoring}. It computes three commonly used evaluation measures MUC~\citep{vilain1995model}, $\text{B}^3$~\citep{bagga1998algorithms}, and $\text{CEAF}_{e}$~\citep{luo2005coreference}. 
We report the ${F}_1$ score of each evaluation measure and their average as the CoNLL score. 

\paragraph{Competing systems.} 
We employed {\sc cort}\footnote{\url{https://github.com/smartschat/cort}, we used version 0.2.4.5.} \citep{martschat2015latent} as our baseline coreference resolution system.
Here, we compare with the original (one best) outputs of {\sc cort}'s latent ranking model\new{, which is the best-performing model implemented in {\sc cort}.}
We consider two rerankers based on \model.
The first reranking method only uses the log probability for \model to sort the
candidate list (\autoref{eq:coref-resolution}). 
The second method uses a linear combination of both log probabilities from \model and the scores from {\sc cort}, where the coefficients were found via grid search with the CoNLL score on the development set. 

\paragraph{Results.} 
The reranked results on the CoNLL 2012 test set are reported in~\autoref{tab:coref}. \new{The numbers of the baseline are higher than the results reported in \citet{martschat2015latent} since the feature set of {\sc cort} was subsequently extended.}
Lines 2 and 3 in \autoref{tab:coref} present the reranked best results. 
As shown in this table, both reranked results give more than 1\% of CoNLL score improvement on the test set over {\sc cort}, which are significant based on an approximate randomization test\footnote{\url{https://github.com/smartschat/art}}.

\begin{table*}
  \centering
  {\small
    \begin{tabular}{@{\extracolsep{0em}}lllll|lll|lll@{\extracolsep{0em}}}
      \toprule
        & & \multicolumn{3}{l}{MUC} & \multicolumn{3}{l}{$\text{B}^3$} & \multicolumn{3}{l}{$\text{CEAF}_{e}$} \\
      \cline{3-5} \cline{6-8} \cline{9-11}
      Model & CoNLL & P & R & $F_1$  & P & R & $F_1$ & P & R & $F_1$\\
      \midrule
      1. Baseline: \textsc{cort}'s one best & 62.93 & 77.15 & 68.67 & 72.66 & 66.00 & 54.92 & 59.95 & 60.07 & 52.76 & 56.18\\
      2. Rerank: \model & {\bf 64.00} & 77.90 & 69.45 & 73.44 & 66.84 & 56.12 & 61.01 & 61.73 & 53.90 & 57.55\\
      3. Rerank: \model + {\sc cort} & {\bf 64.04} & 77.93 & 69.49 & 73.47 & 67.08 & 55.99 & 61.04 & 61.76 & 53.98 & 57.61 \\
      \bottomrule
    \end{tabular}
  }
  \caption{Coreference resolution scores on the CoNLL 2012 test
    set. \textsc{cort} is the best-performing model of \citet{martschat2015latent} with greedy decoding.}
  \label{tab:coref}
\end{table*}

Additional experiments also found that increasing $k$ from 100 to 500 had a
minor effect.
That is because the diversity of each $k$-best list is limited by (i)
the number of entity mentions in the document, (ii) the performance of
the baseline coreference resolution system, and possibly (iii) the
approximate nature of our $k$-best inference procedure.  
We suspect that a stronger baseline
system (such as that of \citealp{clark2016ranking}) could give greater
improvements, if it can be adapted to provide $k$-best lists.
Future work might incorporate the techniques embedded in such
systems into \model.


\subsection{Entity prediction}
\label{subsec:app-ep}


\begin{figure} 
\centering
\begin{tabular}{p{0.45\textwidth}}
  \toprule
  \entone{I} was about to ride \entone{my} \enttwo{bicycle} to the \entthree{park} one day when \entone{I} noticed that the front \entfour{tire} was flat .
  \entone{I} realized that \entone{I} would have to repair \entfour{it} .
  \entone{I} went into \entone{my} \entfive{garage} to get some \entsix{tools} .
  The first thing \entone{I} did was remove the \colorbox{blue!30}{xxxx}\\
  \bottomrule
\end{tabular}
\caption{ A short story on bicycles from the InScript corpus~\citep{modi2017modeling}.
  The entity prediction task requires predicting \colorbox{blue!30}{xxxx} given the preceding text either by choosing a previously mentioned entity or deciding that this is a ``new entity''.
  In this example, the ground-truth prediction is \entfour{tire}.
  For training, \model attempts to predict every entity. While, for testing, 
  it predicts a maximum of 30 entities after the first three sentences, which is 
  consistent with the experimental setup suggested by~\citet{modi2017modeling}.}
\label{tab:entity_prediction_example}
\end{figure}

\paragraph{Task description.}
Based on \citet{modi2017modeling}, we introduce a novel entity prediction task that tries to predict the next entity given the preceding text.
For a given text as in~\autoref{tab:entity_prediction_example}, this task makes a forward prediction based on only the left context. 
This is different from coreference resolution, where both left and right contexts from a given entity mention are used in decoding.
It is also different from language modeling, since this task only requires predicting entities.
Since \model is generative, it can be directly applied to this task.
To predict entities in test data, ${R}_t$ is always given and \model only needs to predict ${E}_t$ when $R_t=1$.

\paragraph{Baselines and human prediction.}
We introduce two baselines in this task: (i) the {\bf always-new} baseline that always predicts ``new entity'';
(ii) a linear classification model using {\bf shallow features} from \citet{modi2017modeling}, including the recency of an entity's last mention and the frequency.
We also compare with the model proposed by~\citet{modi2017modeling}.
Their work assumes that the model has prior knowledge of all the participant types, 
which are
specific to each scenario and fine-grained, e.g., rider in the bicycle narrative,
and predicts participant types for new entities.
This assumption is unrealistic for pure generative models like ours. 
Therefore, we remove this assumption and adapt their prediction results to our formulation by mapping all the predicted entities that have not been mentioned to ``new entity''.
We also compare to the adapted {\bf human prediction} 
used in the InScript corpus. 
For each entity slot, \citet{modi2017modeling} acquired 20 human predictions, and the majority vote was selected. 
More details about human predictions are discussed in~\citep{modi2017modeling}.



\paragraph{Results.}
\tableref{tab:entity_prediction_performance} shows the prediction accuracies.
\model (line 4) significantly outperforms both baselines (line 1 and 2)
and prior work (line 3) ($p \ll 0.01$, paired $t$-test).
The comparison between line 4 and 5 shows our model is even close to the human prediction performance.


\begin{table}
  \centering
  {\small
  \begin{tabular}{p{0.29\textwidth}l}
    \toprule
    & Accuracy (\%)\\
    \midrule
    1. Baseline:  always-new & 31.08 \\
    2. Baseline: shallow features & 45.34 \\[0.3em]
    3. \citet{modi2017modeling} & 62.65 \\[0.3em]
    4. \model & {\bf 74.23} \\[0.3em]
    5. \emph{Human prediction} & 77.35 \\
    \bottomrule
  \end{tabular}
}
  \caption{Entity prediction accuracy on the test set of the InScript corpus. 
  \label{tab:entity_prediction_performance}  }
\end{table}



\section{Related Work}


\paragraph{Rich-context language models.} The originally proposed recurrent neural network language models only capture information within sentences. 
To extend the capacity of RNNLMs, various researchers have
incorporated information beyond sentence boundaries.
Previous work focuses on contextual information from previous sentences~\citep{ji2016document} or discourse relations between adjacent sentences~\citep{ji2016latent},
showing improvements to language modeling and related tasks like coherence evaluation and discourse relation prediction.
In this work, \model adds explicit entity information to the language model, which is another way of adding a memory network for language modeling. 
Unlike the work by~\citet{monz2016recurrent}, where memory blocks are used to store general contextual information for language modeling, \model assigns each memory block specifically to only one entity.

\paragraph{Entity-related models.} 
Two recent approaches to modeling entities in text are closely related to our model. 
The first is the ``reference-aware'' language models proposed by~\citet{yang2016reference}, 
where the referred entities are from either a predefined item list, an
external database, or 
the context from the same document. 
\citet{yang2016reference} present three models, one for each case. 
For modeling a document with entities, they use coreference links to
recover entity clusters, though they only model entity
mentions as containing a single word (an inappropriate assumption, in our view).
Their entity updating method takes the latest hidden state
(similar to $\vec{h}_t$ when $R_t=1$ in our model) as the new
representation of the current entity; no long-term history of the
entity is maintained, just the current local context.
In addition, their language model evaluation assumes that entity information
is provided at test time~(Yang, personal communication), which makes a direct comparison with our model impossible.
Our entity updating scheme is similar to the ``dynamic memory'' method used by~\citet{henaff2016tracking}.
\new{
  Our entity representations are dynamically allocated and updated only when an entity appears up,
  while the EntNet from~\citet{henaff2016tracking} does not model entities and their relationships explicitly. 
  In their model, entity memory blocks are pre-allocated and updated simultaneously in each timestep.
  So there is no dedicated memory block for every entity and no distinction between entity mentions and non-mention words. 
  As a consequence, it is not clear how to use their model for coreference reranking and entity prediction.
}

\paragraph{Coreference resolution.} The hierarchical structure of our
entity generation model is inspired by \citet{haghighi2010coreference}. 
They implemented this idea as a probabillistic graphical model with the distance-dependent Chinese Restaurant Process \citep{pitman1995exchangeable} for entity assignment, while our model is built on a recurrent neural network architecture. 
The reranking method considered in our coreference resolution evaluation
could also be extended with samples from additional coreference
resolution systems, to produce more variety \citep{ng2005machine}. 
The benefit of such a system comes, we believe, from the explicit
tracking of each entity throughout the text, providing entity-specific representations.
In previous work, such information has been added as features \cite{luo2004mention,bjorkelund2014perceptron} or by computing distributed entity representations \cite{wiseman2016global,clark2016improving}. 
Our approach complements these previous methods.

\paragraph{Entity prediction.} 
The entity prediction task discussed in~\S\ref{subsec:app-ep} is based
on work by \citet{modi2017modeling}. 
The main difference is that we do not assume that all entities belong to a previously known set of entity types specified for each narrative scenario.
This task is also closely related to the ``narrative cloze'' task of \citet{chambers2008unsupervised} and the ``story cloze test'' of \citet{Mostafazadeh:ProceedingsOfThe2016NorthAmericanChapter:2016}.
Those studies aim to understand relationships between events, 
while our task focuses on predicting upcoming entity mentions.

\section{Conclusion}
\label{sec:conclusion}

We have presented a neural language model, \model, that defines a
distribution over texts and the mentioned entities.
It provides vector representations for the entities and updates them dynamically in context. 
The dynamic representations are further used to help generate specific entity mentions and the following text. 
This model outperforms strong baselines and prior work on three tasks: language modeling, coreference resolution, and entity prediction. 


\section*{Acknowledgments}
We thank anonymous reviewers for the helpful feedback on this work.
We also thank the members of Noah's ARK and XLab at University of Washington for their valuable comments, particularly  Eunsol Choi for pointing out the InScript corpus. 
This research was supported in part by a University of Washington
Innovation Award, Samsung GRO, NSF grant IIS-1524371, the DARPA CwC program through ARO (W911NF-15-1-0543), and gifts by Google and Facebook.

\bibliography{myref}
\bibliographystyle{emnlp_natbib}

\end{document}